\documentclass[runningheads]{llncs}

\usepackage[english]{babel}

\usepackage{amsmath}
\usepackage{graphicx}
\usepackage[colorlinks=true, allcolors=blue]{hyperref}

\usepackage{lipsum}  
\usepackage{xargs} 
\usepackage{booktabs}
\usepackage{multirow}
\usepackage{multicol}
\usepackage{tikz} % rounded tables
\usepackage{sectsty}
\usepackage[pdftex,dvipsnames,table,xcdraw]{}
\usepackage{colortbl}

\usepackage{xspace}
\usepackage{authblk}

\usepackage[colorinlistoftodos,prependcaption,textsize=tiny]{todonotes}
\newcommandx{\unsure}[2][1=]{\todo[linecolor=red,backgroundcolor=red!25,bordercolor=red,#1]{#2}}
\newcommandx{\change}[2][1=]{\todo[linecolor=blue,backgroundcolor=blue!25,bordercolor=blue,#1]{#2}}
\newcommandx{\info}[2][1=]{\todo[linecolor=OliveGreen,backgroundcolor=OliveGreen!25,bordercolor=OliveGreen,#1]{#2}}
\newcommandx{\improvement}[2][1=]{\todo[linecolor=violet,backgroundcolor=violet!25,bordercolor=violet,#1]{#2}}
\newcommandx{\thiswillnotshow}[2][1=]{\todo[disable,#1]{#2}}

\newcommand{\model}{Sabiá\xspace}
\newcommand{\evalset}{Poeta\xspace}

\title{\includegraphics[width=0.06\textwidth]{sabia.png} \model: Portuguese Large Language Models
}
\titlerunning{\model: Portuguese Large Language Models}
\author{Ramon Pires \and
Hugo Abonizio \and
Thales Sales Almeida \and
Rodrigo Nogueira
}
\authorrunning{R. Pires et al.}

\institute{Maritaca AI \\
\email{\{ramon,hugo,thales,rodrigo\}@maritaca.ai}}

\begin{document}
\maketitle

\begin{abstract}

As the capabilities of language models continue to advance, it is conceivable that ``one-size-fits-all" model will remain as the main paradigm. For instance, given the vast number of languages worldwide, many of which are low-resource, the prevalent practice is to pretrain a single model on multiple languages.
In this paper, we add to the growing body of evidence that challenges this practice, demonstrating that monolingual pretraining on the target language significantly improves models already extensively trained on diverse corpora.
More specifically, we further pretrain GPT-J and LLaMA models on Portuguese texts using 3\% or less of their original pretraining budget. Few-shot evaluations on \textit{\evalset}, a suite of 14 Portuguese datasets, reveal that our models outperform English-centric and multilingual counterparts by a significant margin. Our best model, \model-65B, performs on par with GPT-3.5-turbo. By evaluating on datasets originally conceived in the target language as well as translated ones, we study the impact of language-specific pretraining in terms of 1) capturing linguistic nuances and structures inherent to the target language, and 2) enriching the model's knowledge about a domain or culture. Our results indicate that most benefits stem from the domain-specific knowledge acquired through monolingual pretraining. Finally, we show that our Portuguese model has a lower performance in English tasks, thereby substantiating the inherent compromise in refining models for specific domains. Sabiá-7B is avaliable at HuggingFace: \url{https://huggingface.co/maritaca-ai/sabia-7b}

%\keywords{First keyword  \and Second keyword \and Another keyword.}
\end{abstract}

\section{Introduction}
\label{sec:introduction}

Language Models have revolutionized the field of natural language processing with their exceptional ability to perform tasks with minimal supervision. Although primarily pretrained on English-centric corpora, the models have shown impressive multilingual capabilities~\cite{chowdhery2022palm}. Given the abundance of languages worldwide, the majority of which are low-resource, it has become a common practice to pretrain single models on multiple languages simultaneously. Models like XLM-R~\cite{conneau2020unsupervised}, mBART~\cite{liu2020multilingual}, mT5~\cite{xue2020mt5}, and BLOOM~\cite{scao2022bloom} exemplify this approach.

Despite the success of these multilingual models, we argue that they may not be the optimal approach for capturing the cultural and knowledge richness inherent in individual languages.
When a moderately-sized language-specific corpus is available, continued pretraining could integrate the missing knowledge into the model, enhancing its performance on targeted tasks.
To test this hypothesis, we extend the pretraining of English-centric models using Portuguese corpora and evaluate their performance on an extensive range of Portuguese datasets employing a few-shot learning approach.  Our results indicate that, even for models trained beyond the recommendations by Hoffmann et al~\cite{hoffmann2022training}, this additional pretraining considerably improves performance compared to multilingual models.

We evaluate our models on datasets comprising texts originally created by native Brazilian Portuguese speakers, as well as datasets translated from English to Portuguese. We observe improvements across all datasets due to the Portuguese pretraining, with the gains being particularly pronounced for datasets created by Brazilian speakers. One of the largest improvements was observed on the ENEM dataset~\cite{silveira2018enem}, which is derived from entrance exams used by Brazilian universities and requires extensive knowledge of the country's history, geography, and literature. 
This result provides evidence that the major contribution of our language-specific pretraining is to inject domain-specific knowledge about a particular culture as opposed to solely enhancing language proficiency.

\section{Related Work}
\label{sec:related-work}

The success of multilingual pretraining has been well-documented in the literature, with models such as ByT5~\cite{xue2022byt5}, mT5~\cite{xue2020mt5}, XLM-R~\cite{conneau2020unsupervised}, XGLM~\cite{lin2022few} and mGPT~\cite{shliazhko2022mgpt} paving the way for more inclusive language understanding and generation by leveraging shared knowledge across multiple languages. However, there are limitations to this approach.

BLOOM, a 175B-parameter model pretrained on 46 languages, performs worse on English tasks compared to OPT~\cite{zhang2022opt}, a similarly sized model pretrained on English-centric corpora using comparable computational resources and data size. We conjecture that BLOOM's underperformance may be attributed to its relatively limited exposure to English tokens during the pretraining phase. Consequently, this observation suggests that monolingual pretraining could offer supplementary advantages.

In support of this hypothesis, models with hundreds of millions of parameters pretrained on monolingual texts have demonstrated gains over multilingual counterparts~\cite{CaneteCFP2020,souza2020bertimbau,carmo2020ptt5,sarti2022it5,martin2020camembert,le2020flaubert,chan2020germans,arabert,lee2021korealbert,phobert,kalyan2021ammus}. Additionally, research has indicated that language adaptation is beneficial even for low-resource languages~\cite{ogueji2021small,ebrahimi2021adapt,bhattacharjee2022banglabert,yong2022bloom+}.
However, there is a limited number of published research articles with comprehensive evaluations of the benefits of continued pretraining at the multi-billion-parameter scale~\cite{BERTIN-GPT,zeng2022glm,kim2021changes}. Through this study, we contribute to the literature by demonstrating the effectiveness of continued language-specific pretraining for Portuguese language models up to the 65B-parameter scale.

The question concerning whether it is advantageous to train models for specific languages is closely associated with the question of whether it is beneficial to train models for particular domains of knowledge. Recent studies, such as Minerva~\cite{lewkowycz2022solving} and Galactica~\cite{taylor2022galactica}, have shown that domain-specific pretraining can lead to significant improvements, even with a smaller pretraining corpus compared to large-scale, general-purpose pretraining corpora. 
Analogously, Fu et al.~\cite{fu2023specializing} demonstrated the feasibility of specializing smaller models to perform multi-step reasoning, a capability typically exclusive to models with at least 50B parameters, at the expense of diminished performance in other, more general tasks.

Pretraining with a combination of general and domain-specific corpora can potentially enhance performance in specialized tasks without compromising effectiveness in general-purpose tasks, albeit at the cost of increased computational demands. For example, BloombergGPT~\cite{wu2023bloomberggpt}, a 50B-parameter model pretrained on heterogeneous corpus in which more than half of texts are from the financial domain, exhibits comparable performance to OPT-66B in general tasks. However, BloombergGPT's pretraining dataset is three times larger, and consequently used more computational resources.

Rather than pursuing a single model that performs well across multiple domains, Gururangan et al.~\cite{gururangan2023scaling} propose an alternative approach: using multiple expert models, each trained on a domain-specific subset within a broader, diverse dataset, to function as a single general-purpose model. Their models outperform dense ones across various domain-specific tasks, at the expense of an increased parameter count, consequently leading to larger memory requirements for efficient inference.\footnote{To serve their ensemble with a low latency, the weights for each expert must be kept in GPU memory.}

\section{Methodology}

In this section, we outline the pretraining data and training details used to build our models, including data sources, preprocessing techniques, architectures, hyperparameters, and optimization methods.

\subsection{Pretraining Data}
\label{sec:pretrained-data}

The pretraining data is derived from the Portuguese subset of the ClueWeb 2022 dataset~\cite{overwijk2022clueweb22,overwijk2022clueweb22_arxiv}. To increase the datasets's quality, we apply the quality filters from MassiveText~\cite{rae2021scaling}, modifying them to accommodate the specific requirements of the Portuguese language. We normalize the text with \textit{ftfy}%~\cite{speer-2019-ftfy}
\footnote{\emph{ftfy} normalization fixes \textit{mojibakes} and remove remnant HTML tags.}, convert wikitexts into human-readable texts, and exclude documents containing less than 200 unique tokens. 

These quality filters are primarily designed for web pages and may not seamlessly transfer to other domains. There is potential for improvement by employing more automated methods; however, this study did not explore such approaches due to the resource-intensive nature of pretraining experiments.

Following the cleaning process, all documents are concatenated using an end-of-sequence token as a separator, and then tokenized. The GPT-J tokenizer, which is identical to the GPT-2 tokenizer~\cite{radford2019language}, produces 7.8 billion tokens, while the LLaMA tokenizer produces 7.3 billion tokens. 
The discrepancy in the total number of tokens is primarily due to the different tokenization strategies each model employs, byte-level BPE and BPE based on sentencepiece~\cite{kudo2018sentencepiece}, respectively along with the variation of the vocabularies used by each tokenizer.

We extended the training of three models — LLaMA 7B and 65B~\cite{touvron2023LLaMA} as well as GPT-J~\cite{gpt-j} — originally trained on English-centric corpora, on Portuguese texts; these further pretrained models from LLaMA are denoted as \model, while the one derived from GPT-J is referred to as \model-J.\footnote{\model is a tribute to the eponymous bird, renowned for its diverse and intricate vocalizations.}

\subsection{\model models}

The LLaMA 7B and 65B models are decoder-only Transformer models~\cite{vaswani2017attention} with a similar architecture to PALM's~\cite{chowdhery2022palm}. The models were trained using a causal language modeling objective on a massive dataset sourced from webpages, code, books, and scientific papers. The 7B model was trained on 1 trillion tokens and the 65B model was trained on 1.4 trillion tokens. % The majority of the corpus consists of English text; however, it also contains Portuguese text, with the precise quantity not specified by the authors.
While the majority of the corpus is in English, it also includes an unspecified amount of Portuguese text.

Starting from the LLaMA weights, we train the \model models on our Portuguese dataset (see Section~\ref{sec:pretrained-data}) using the \texttt{t5x} and \texttt{seqio} frameworks~\cite{roberts2022scaling}. Adhering closely to the hyperparameters used by PALM, we use the AdaFactor optimizer~\cite{shazeer2018adafactor} without factorization, a first-order momentum $\beta_1 = 0.9$, and a second-order momentum $\beta_2 = 1 - k^{-0.8}$, where $k$ represents the step number. We apply global norm clipping at 1.0 and dynamic weight decay of $lr^2$, with $lr$ denoting the current learning rate.

Besides the standard causal language modeling loss, we use an auxiliary loss of $10^{-4} \log^2 (\sum_i e^{z_i})$, where $z$ are the logits, to decrease the likelihood of loss spikes at the 65B-parameter scale. The learning rate is linearly increased from 0 to 1e-3 over the initial 1,000 steps, followed by a constant learning rate of 1e-3 for an additional 9,000 steps.

The models were trained on a TPU v2-512, using batches of 512 sequences, each containing 2048 tokens. We utilized gradient checkpointing, also known as rematerialization, to enable the use of larger batches, thereby increasing TPU utilization. For the 7B model, this configuration results in a throughput of 124,000 tokens/sec, corresponding to a Model FLOPs Utilization (MFU)~\cite{chowdhery2022palm} of 45.2\%, excluding the self-attention operations. For the 65B model, we achieve a throughput of 14,000 tokens/sec, resulting in an MFU of 47.4\%.

The resulting models were trained on a total of 10.4 billion tokens, or 1.52 epochs of the Portuguese dataset. This equals to 10,000 training steps.
We noticed improvements in few-shot tasks beyond one epoch, which corroborates results from Taylor et al.~\cite{taylor2022galactica}. However, due to the high costs of pretraining, we did not continue training.\footnote{Considering the on-demand pricing of 384 USD per hour for a TPU v2-512, pretraining \model-7B and \model-65B costs approximately 9,000 and 80,000 USD, respectively.}

\subsection{\model-J}

The GPT-J model is a 6B-parameter decoder-only Transformer model whose architecture and training hyperparameters closely follow GPT-3 6.7B. The main differences reside on computing the MLP and self-attention in parallel, applying attention head with dimension 256 (twice larger than GPT-3 6.7B), and using Rotary Positional Embedding (RoPE)~\cite{rope-paper}. GPT-J was trained on 400B tokens from The Pile dataset~\cite{gao2020pile}, whose 97.4\% tokens are in English.

We begin training \model-J from the released GPT-J checkpoint,\footnote{\url{https://huggingface.co/EleutherAI/gpt-j-6b}} using the \texttt{mesh-transformer-jax} framework~\cite{mesh-transformer-jax} and AdamW optimizer~\cite{loshchilov2018decoupled} with a weight decay of 0.1. We start the pretraining by warming up the learning rate until 1.2e-5 over 13,500 steps, followed by a cosine annealing decay during 135,518 steps until the end learning rate of 2.4e-6, and kept it constant from there on.
We train on a TPU v3-8 using an effective batch size of 32 sequences of 2048 tokens. This results in a throughput of 5,200 tokens/sec, corresponding to a MFU of 44.5\% without self-attention. The model was trained for 18 days on 7.8B tokens, or one epoch of the Portuguese dataset.\footnote{Due to constraints in our hardware budget, this model was trained with fewer tokens compared to \model.}

\section{Evaluation on \evalset}
\label{sec:evaluation}

We evaluate the \model models on the Portuguese Evaluation Tasks (\evalset) benchmark, which comprises 14 downstream NLP datasets in Portuguese: ASSIN 2 RTE and STS~\cite{real2020assin}, ENEM Challenge~\cite{silveira2018enem}, ENEM 2022~\cite{nunes2023evaluating}, FaQuAD~\cite{sayama2019faquad}, TweetSentBr~\cite{brum2018building}, AG News~\cite{Zhang2015CharacterlevelCN}, IMDB~\cite{maas-etal-2011-learning}, MASSIVE~\cite{fitzgerald2022massive}, MKQA~\cite{longpre2021mkqa}, BoolQ~\cite{clark-etal-2019-boolq}, SST2~\cite{socher-etal-2013-recursive}, WSC~\cite{melo2019winograd}, and BLUEX~\cite{almeida2023bluex}.
Half of them (ASSIN 2 RTE and STS, BLUEX, ENEM Challenge, ENEM 2022, FaQuAD, and TweetSentBr) were originally written in Portuguese, and the remaining ones were either manually or automatically translated into Portuguese from their originals in English. We refer to the first group as ``Native'' datasets and the second group as ``Translated'' datasets.\footnote{The MASSIVE dataset underwent manual translation and localization; however, given that the original text was composed in English, it has been categorized as a translated dataset.} %More details about the datasets can be found in Appendix~\ref{app:datasets}.

The models were evaluated in a few-shot manner using the maximum number of examples that fits into a 2048-token context for each task.
We used the \emph{GPT-2} tokenizer as a reference because it results in more tokens. This allowed us to comfortably fit prompts tokenized with other tokenizers.

To evaluate the models, we manually select a set of few-shot examples for each dataset on \evalset. Depending on the dataset, these examples are balanced by class (except for FaQuAD, BLUEX, ENEM Challenge, ENEM 2022, MKQA, and WSC). For each test example, the prompts are built with the selected few-shot examples in alternating order.
%We also explore different alternatives to select few-shot examples in an ablation study (see Section~\ref{sec:ablation_prompt_modes}).
Each task on \evalset has a particular instruction that is placed at the beginning of the prompt.

Following Srivastava et al~\cite{srivastava2022beyond}, we adopt the Normalized Preferred Metric (NPM) as our primary evaluation measure:

\begin{equation}
    \texttt{NPM} = \frac{1}{N} \sum_{i=1}^N 100 \times \frac{\texttt{[raw preferred metric]}_i - \texttt{[random score]}_i}{\texttt{[high score]}_i - \texttt{[random score]}_i}
\end{equation}

\noindent where $N$ is the number of evaluation datasets, $\texttt{[raw preferred metric]}_i$ is the score obtained by the model on the $i$-th dataset, $\texttt{[random score]}_i$ is the score of a random model (e.g., 50\% for a binary classification task) and $\texttt{[high score]}_i$ is the highest possible score on that dataset, which is either 1 or 100. The preferred metric and random score for each dataset are presented in Table~\ref{tab:dataset_summary}.
The rationale behind employing NPM rather than a straightforward average across all datasets is to mitigate the undue influence of datasets with inherently high scores, such as binary classification datasets, which could otherwise outweigh datasets characterized by lower scores.

\begin{table}[!t]
\centering\resizebox{1.0\textwidth}{!}{
\begin{tabular}{@{}l|cccccccc@{}}
\toprule
& & \textbf{Preferred} & \textbf{Rand.} & & \textbf{Avg Len} & \textbf{Num} & \textbf{Num} & \textbf{Num}\\
\textbf{Dataset} & \textbf{Type} & \textbf{Metric}  &  \textbf{Score} & \textbf{Transl.} & \textbf{(chars)} & \textbf{Train} & \textbf{Test} & \textbf{Few-shot}\\
\midrule
AG News & Multiclass classification (4) & Accuracy & 25 & Yes & 282.34 & 120,000 (110,953) & 7,600 & 12 \\ % \footnote{excluding training examples that exceed 298 chars}
ASSIN 2 RTE & Binary classification & F1 & 50 & No & 139.99 & 6,500 & 2448 & 18 \\
ASSIN 2 STS & Regression & Pearson & 0 & No & 139.99 & 6,500 & 2448 & 15 \\
BLUEX & Multiple choice (4) & Accuracy & 25 & No & 1,228.08 & - & 178 & 1 \\
BoolQ & Binary classification & Accuracy & 50 & Yes & 562.30 & 9,427 (7,015) & 3,270 & 4 \\ % \footnote{excluding training examples that exceed 749 chars, and truncating test examples at 1,169 chars}
ENEM Challenge & Multiple choice (5) & Accuracy & 20 & No & 1,286.68 & - & 916 & 1 \\
ENEM 2022 & Multiple choice (5) & Accuracy & 20 & No & 1,170.24 & - & 118 & 1 \\
FaQuAD & Extractive QA & F1 & 0 & No & 1,056.47 & - & 63 & 4\\
IMDB & Binary classification & Accuracy & 50 & Yes & 1,114.56 & 25,000 (18,613) & 25,000 & 2 \\ % \footnote{excluding training examples that exceed 1,544 chars, and truncating test examples at 2,095 chars}
MASSIVE & Multiclass classification (18) & F1-macro & 0.58 & Yes & 68.35 & 11,514 & 2,974 & 36 \\
MKQA & Extractive QA & F1 & 0 & Yes & 80.32 & - & 10,000 (6,758) & 40 \\  % \footnote{excluding examples of unanswerable and long-answer types} 
SST2 & Binary classification & Accuracy & 50 & Yes & 84.19 & 67,349 & 872 & 34 \\
TweetSentBR & Multiclass classification (3) & F1-macro & 32.4 & No & 93.32 & 12,990 & 2010 & 30 \\
WSC & Binary classification & Accuracy & 50 & Yes & 102.15 & - & 285 & 18 \\
\bottomrule
\end{tabular}
}
\caption{A summary of the datasets constituting the \evalset benchmark.}
\label{tab:dataset_summary}
\end{table}

\section{Results}

The main results can be found in Table~\ref{tab:main_results}.
Models such as BLOOMZ, XGLM and Bertin-GPT struggled to generate answers in Portuguese. To address this issue, we adopted an approach akin to that used by the XGLM authors: by calculating the likelihood of each candidate answer string based on the input text and subsequently selecting the class with the highest probability. For FaQuAD, the only dataset in the benchmark without predetermined candidate answers, we allowed the models to generate answers in their original format.

We observe that the LLaMA baselines significantly outperform models of equivalent size trained with fewer tokens, such as Galactica and OPT. Furthermore, despite being trained on English-centric corpora, LLaMA-7B surpasses multilingual BLOOM and XGLM of similar sizes. The \model models demonstrate considerable improvement in NPM compared to their respective baseline models. These NPM gains are more substantial for the smaller \model-J and \model-7B models. Notably, \model-65B marginally outperforms OpenAI's GPT-3.5-turbo, which serves as the base model for ChatGPT.

\begin{table}[!htb]
\scriptsize
\caption{Few-shot NPM results on the \evalset benchmark.}
\label{tab:main_results}
\centering
\begin{tabular}{@{}lccc@{}}
\toprule
 & \textbf{Native} & \textbf{Translated} & \textbf{All}\\
\midrule
GALACTICA-6.7B & 2.2 & 13.6 & 7.9\\
OPT-6.7B & 5.3 & 39.7 & 22.5 \\
OPT-66B & 16.4 & 47.1 & 31.7 \\
BERTIN-GPT & 5.8 & 42.5 & 24.2\\
BLOOM-7.1B & 10.6 & 44.2 & 27.4\\
BLOOMZ-7.1B & 18.3 & 44.7 & 31.5\\
XGLM-7.5B & 14.0 & 46.9 & 30.4 \\
GPT-3.5-turbo & 67.9 & 66.0 & 67.0 \\
GPT-4 & 78.8 & 82.5 & 80.6\\
 \midrule 
GPT-J & 10.2 & 33.9 & 22.0 \\
\model-J & 25.0 & 43.1 & 34.0\\
\midrule 
LLaMA-7B & 20.2 & 45.8 & 33.0 \\
\model-7B & 43.4 & 53.6 & 48.5 \\
\midrule 
LLaMA-65B & 59.1 & 68.4 & 63.7 \\
\model-65B & 69.2 & 69.6 & 69.4 \\
\bottomrule
\end{tabular}
\end{table}

Through our Portuguese pretraining, we observed that the improvement in NPM was higher in native datasets than that in translated datasets. For \model-65B, improvements over LLaMA-65B were mostly from the native subset. We hypothesize that this is due to the ``mechanistic'' nature of translated datasets: since they were translated from English, the baseline model already possesses the knowledge needed to solve them and gains little from learning the linguistic, syntactic, and grammatical knowledge of the target language. For instance, to answer the question ``\textit{does p o box come before street address}'' (BoolQ dataset), the model gains little from additional pretraining on a Portuguese corpus as it is unlikely that the corpus would provide new information regarding the formatting of US mailing addresses that the model has not already encountered during its initial English-centric pretraining.
Conversely, language-specific pretraining introduces the specific knowledge required to solve tasks in the native subset.

Although GPT-J exhibited lower few-shot performance in English tasks relative to LLaMA, we use it in this study to illustrate that not only highly optimized models like LLaMA can benefit from extended pretraining. We chose not to use BLOOM-7.1B as our initial checkpoint for pretraining due to its inferior performance compared to GPT-J in preliminary few-shot experiments on three Portuguese datasets. However, we later discovered that its performance on \evalset surpassed GPT-J's. Nonetheless, BLOOM still exhibits lower performance compared to LLaMA.

Analogous to \model-J, BERTIN-GPT is a model pretrained on Spanish text starting from the GPT-J weights. Since Spanish and Portuguese are similar languages, it is reasonable to expect that BERTIN-GPT would perform better than its baseline model.
Nevertheless, the observed NPM for BERTIN-GPT is only slightly higher than GPT-J's.

A noteworthy comparison involves Galactica, a model pretrained on scientific text, predominantly in English, and a similarly-sized OPT model, which utilized comparable pretraining compute but was pretrained on a larger and more diverse English-centric corpus. In their study, the authors demonstrate that Galactica performs on par with OPT on English tasks and largely outperforms OPT on scientific-related tasks. Conversely, OPT significantly outperforms Galactica in Portuguese tasks. This result underscores the trade-offs associated with domain-specific specialization, which often entails diminished performance in other tasks.

BLOOMZ~\cite{muennighoff2022crosslingual}, a multilingual instruction-tuned model, demonstrated superior performance compared to its baseline BLOOM model, rivaling LLaMA of equivalent size.\footnote{This model was used in the experiments: \url{https://huggingface.co/bigscience/bloomz-7b1-mt}} Nevertheless, our approach of pretraining in Portuguese appears to yield superior results, as \model-J surpasses BLOOMZ despite originating from a lower-performing baseline model. We envision continued pretraining and instruction tuning as complementary techniques to be combined in future research.

\subsection{Results per Dataset}

Table~\ref{tab:results_per_dataset} presents the results per \evalset dataset for \model models, their baselines, and for the supervised state-of-the-art. The SOTA results reported for the translated datasets were obtained using their original English versions~\cite{xlnet2019yang,zoph2022designing,raffel2020exploring,sakaguchi2021winogrande}. Since the \evalset benchmark excludes unanswerable examples of the MKQA dataset, we decided not to include the SOTA result for this dataset.

In more challenging datasets, such as ENEM Challenge, ENEM 2022, and BLUEX, which are derived from admission exams to Brazilian universities, we see the most significant gains due to language-specific pretraining. Substantial improvements are also observed in TweetSentBr, a dataset containing tweets with an abundance of slang and references to Brazilian popular culture. We hypothesize that this pretraining imparts specific knowledge about the country's culture, literature, and geography that is less frequently encountered and learned during the original pretraining with more diverse texts.

\begin{table}[!htb]
\caption{Results per dataset. $^1$\cite{rosa2021cost}; $^2$\cite{Chaves_Rodrigues_eplm_2023}; $^3$\cite{moraes2021cost}; $^4$\cite{barros2021employing}; $^5$\cite{xlnet2019yang}; $^6$\cite{zoph2022designing}; $^7$\cite{raffel2020exploring}; $^8$\cite{sakaguchi2021winogrande}.}
\label{tab:results_per_dataset}
\resizebox{\columnwidth}{!}{%
\begin{tabular}{@{}lccccccccccccccc@{}}
\toprule
                          & \multicolumn{1}{l}{} & \multicolumn{7}{c}{\textbf{Native}}                            & \multicolumn{7}{c}{\textbf{Translated}}                        \\  \cmidrule(l){3-9} \cmidrule(l){10-16}  %\midrule
 &
  Avg &
  \begin{tabular}[c]{@{}c@{}}ASSIN 2 RTE\\ (F1)\end{tabular} &
  \begin{tabular}[c]{@{}c@{}}ASSIN 2 STS\\ (Pearson)\end{tabular} &
  \begin{tabular}[c]{@{}c@{}}BLUEX\\ (Acc)\end{tabular} &
  \begin{tabular}[c]{@{}c@{}}ENEM\\ (Acc)\end{tabular} &
  \begin{tabular}[c]{@{}c@{}}ENEM 2022\\ (Acc)\end{tabular} &
  \begin{tabular}[c]{@{}c@{}}FaQuAD\\ (F1)\end{tabular} &
  \begin{tabular}[c]{@{}c@{}}TweetSentBr\\ (F1-macro)\end{tabular} &
  \begin{tabular}[c]{@{}c@{}}AG News\\ (Acc)\end{tabular} &
  \begin{tabular}[c]{@{}c@{}}BoolQ\\ (Acc)\end{tabular} &
  \begin{tabular}[c]{@{}c@{}}IMDB\\ (Acc)\end{tabular} &
  \begin{tabular}[c]{@{}c@{}}MASSIVE\\ (F1-macro)\end{tabular} &
  \begin{tabular}[c]{@{}c@{}}MKQA\\ (F1)\end{tabular} &
  \begin{tabular}[c]{@{}c@{}}SST2\\ (Acc)\end{tabular} &
  \begin{tabular}[c]{@{}c@{}}WSC\\ (Acc)\end{tabular} \\  \midrule  %\cmidrule(l){3-9} \cmidrule(l){10-16} 
SOTA supervised           & -                     & 92.07$^1$ & 86.00$^2$ & -     & -     & -     & 82.40$^3$ & 77.27$^4$ & 95.55$^5$ & 92.40$^6$ & 96.21$^5$ & -     & - & 97.50$^7$ & 90.10$^8$ \\
GPT-4                     & 84.99                & 90.96 & 77.58 & 76.40 & 92.00 & 79.66 & 84.74 & 82.40 & 93.50 & 86.50 & 97.00 & 83.30 & 55.67 & 97.50 & 92.63 \\
GPT-3.5-turbo             & 76.08                & 88.28 & 66.41 & 60.11 & 80.57 & 75.42 & 78.28 & 74.39 & 87.71 & 71.43 & 84.86 & 84.19 & 44.92 & 91.71 & 76.84 \\
Galactica-6.7B            & 34.11                & 34.92 & 11.63 & 28.65 & 20.74 & 22.88 & 40.16 & 21.98 & 38.33 & 57.13 & 51.08 & 35.62 & 3.01 & 62.27 & 49.12       \\
Bertin-GPT-6B             & 45.18                & 33.24 & 6.23 & 22.47 & 20.52 & 22.03 & 64.00 & 35.52 & 82.44 & 44.25 & 87.66 & 55.46 & 15.56 & 81.08 & 62.11 \\
OPT-6.7B                  & 43.35                & 43.33 & 21.35 & 24.16 & 19.87 & 20.34 & 56.45 & 14.37 & 55.67 & 61.31 & 90.42 & 51.84 & 13.64 & 86.47 & 47.72 \\
OPT-66B                   & 49.68                & 65.66 & 7.88  & 29.78 & 20.41 & 17.80 & 71.12 & 32.54 & 81.87 & 58.75 & 92.66 & 61.64 & 21.17 & 87.50 & 46.67 \\
BLOOM-7.1B                & 47.01                & 50.32 & 12.16 & 25.84 & 20.85 & 17.08 & 72.67 & 25.12 & 79.48 & 60.43 & 89.60 & 56.23 & 15.72 & 83.83 & 48.77 \\
BLOOMZ-7.1B               & 50.94                & 33.57 & 24.50 & 34.27 & 28.38 & 27.12 & 79.90 & 50.36 & 83.82 & 38.23 & 93.80 & 55.31 & 12.36 & 86.93 & 64.56 \\
XGLM-7.5B                 & 48.79                & 53.75 & 15.07 & 24.16 & 19.10 & 19.49 & 44.84 & 63.23 & 77.47 & 49.76 & 91.46 & 59.74 & 13.72 & 89.11 & 62.11 \\
\midrule
GPT-J                     & 43.51                & 54.88 & 17.86 & 24.72 & 20.85 & 20.34 & 59.52 & 20.98 & 64.15 & 48.75 & 72.68 & 55.67 & 10.69 & 83.94 & 54.04 \\
\model-J                  & 52.84                & 35.49 & 22.97 & 39.89 & 39.41 & 36.44 & 69.28 & 64.16 & 84.30 & 51.53 & 90.86 & 58.82 & 13.84 & 87.16 & 45.61 \\
\midrule
LLAMA-7B                  & 51.30                & 56.82 & 7.39  & 32.02 & 29.04 & 23.73 & 77.38 & 44.19 & 76.94 & 57.37 & 86.92 & 59.90 & 30.08 & 88.76 & 47.72 \\
\model-7B                 & 62.43                & 64.87 & 13.63 & 47.75 & 60.59 & 60.17 & 77.43 & 67.17 & 83.28 & 64.07 & 92.70 & 68.95 & 31.98 & 90.60 & 50.88 \\
\midrule
LLAMA-65B                 & 73.84                & 74.98 & 62.85 & 53.93 & 75.00 & 62.71 & 87.25 & 68.05 & 88.01 & 73.12 & 94.98 & 78.71 & 48.34 & 94.27 & 71.58 \\
\model-65B                & 77.65                & 88.07 & 63.29 & 57.87 & 90.39 & 72.03 & 88.47 & 72.91 & 88.34 & 75.96 & 92.76 & 79.41 & 49.47 & 93.43 & 74.74 \\ \bottomrule
\end{tabular}%
}
\end{table}

Certain capabilities only emerge at scale, as evidenced by \cite{wei2022emergent}. For example, 6-7B models perform close to the random baseline in datasets such as ASSIN 2 RTE and STS, and WSC. However, at the 65B scale, we observe substantial improvements, approaching or surpassing state-of-the-art supervised models on the ASSIN 2 RTE and FaQuAD datasets.

GPT-4~\cite{openai2023gpt4} results indicate that there is still room for improvement for \model-65B in the majority of the datasets evaluated in this work. Nevertheless, \model-65B performs on par with GPT-4 in datasets such as ASSIN 2 RTE, ENEM Challenge, and FaQuAD.

\subsection{Data Contamination}
The pretraining data for \model models were collected up until February 2022. Since ENEM 2022 was publicly released in November 2022, the model could not have access to the answers for the questions present within its pretraining data. Consequently, the improvements observed at least for ENEM 2022, which were higher than the average of the datasets, cannot be attributed to data contamination. However, for the other datasets, the possibility of data contamination cannot be ruled out.

\subsection{Ablation: English datasets}
\label{sec:ablation_english_tasks}

In this ablation study, we investigate the potential impact of Portuguese pretraining on the performance of the model in English datasets.
We evaluated the LLaMA-7B and the \model-7B models in English multiple-choice tasks.
For simplicity, we employed a few-shot evaluation setup with 10 randomly selected examples (dynamic-sampled prompt).
Importantly, we did not incorporate any descriptions or include Portuguese keywords to delimit the few-shot examples.
We also restricted all the datasets to 350 test examples.

Following LLaMA's~\cite{touvron2023LLaMA} approach, given the provided context, we select the answer with the highest likelihood normalized by the number of characters. The results in Table~\ref{tab:results_english} indicate that the \model-7B model exhibits a slightly reduced performance in English tasks compared to the baseline. 
This result corroborates our premise that model specialization invariably entails a balancing act, where improvements in one domain frequently coincide with degradation in another.

% ## Number of test examples per dataset
% # winogrande    = 1767
% # hellaswag     = 10003
% # ARC-Easy      = 2376
% # ARC-Challenge = 1172
% # piqa          = 3074
% # openbookqa    = 500

% # FROM LLAMA
% We evaluate LLaMA on free-form generation tasks and multiple choice tasks. In the multiple choice tasks, the objective is to select the most appropriate completion among a set of given options, based on a provided context. We select the completion with the highest likelihood given the provided context. We follow Gao et al. (2021) and use the likelihood normalized by the number of characters in the completion, except for certain datasets (OpenBookQA, BoolQ), for which we follow Brown et al. (2020), and select a completion based on the likelihood normalized by the likelihood of the completion given “Answer:” as context: P(completion|context)/P(completion|“Answer:”).

% Restringimos para datasets implementados no lm-eval, todas usando loglikelihood requests.
% Rodamos testes em no máximo 350 exemplos (limit=350).
% Não fizemos uma seleção manual de exemplos few-shot, mas selecionamos aleatoriamente (dynamic-random).
% Usamos 10 exemplos few-shot.

\begin{table}[!htb]
\centering
\caption{Results in English datasets.}
\label{tab:results_english}
\begin{tabular}{@{}lccccccc@{}}
\toprule
          & \textbf{PIQA} & \textbf{HellaSwag} & \textbf{WinoGrande} & \textbf{ARC-e} & \textbf{ARC-c} & \textbf{OBQA} & \textit{\textbf{NPM}} \\ \midrule
LLaMA-7B  & 83.43         & 77.43              & 74.29               & 69.43          & 48.86          & 43.14         & 50.10                     \\
\model-7B & 80.86         & 75.71              & 72.29               & 72.86          & 50.00          & 42.29         & 49.02                     \\ \bottomrule
\end{tabular}
\end{table}

\vspace{-0.5cm}
\section{Limitations}

Owing to the financial constraints associated with pretraining and, more significantly, the manual labor involved in collecting and curating evaluation datasets, experiments were conducted exclusively in Portuguese. Given that our models started pretraining from English-pretrained models and that Portuguese and English exhibit relatively close linguistic proximity, we anticipate that other researchers conducting further pretraining on languages closely related to English will observe comparable improvements in their target tasks. However, determining whether the benefits of this method persist for languages more distant from English remains an open research question.

Portuguese is a language with an abundance of high-quality web-based texts. Thus, the gains observed with the proposed method may not necessarily extend to low-resource languages with limited availability of quality texts. In such cases, parameter-efficient methods~\cite{houlsby2019parameter,pfeiffer2020mad,pfeiffer2021adapterfusion} could be advantageous, as evidenced by Yong et al.~\cite{yong2022bloom+}. We did not use these techniques in this study due to the training costs, which are approximately equivalent to training the entire model.\footnote{Although parameter-efficient methods adjust only a fraction of the weights, they use only marginally fewer training FLOPs, as activations and gradients are computed for the entire model. For instance, LoRA~\cite{hu2022lora}, a parameter-efficient method, improves training throughput of a GPT-3 175B model by only nearly 32\%.}

\section{Conclusion}
\label{sec:conclusion}

In this study, we contributed to the expanding body of scientific evidence that specializing models for individual languages leads to improvements, even when the baseline model is large and extensively trained. We achieved this for the Portuguese language utilizing a near state-of-the-art model with 65 billion parameters.
Given the relatively low pretraining cost and significant performance gains observed, we foresee a future landscape consisting of a diverse array of models, each tailored to a specific domain, rather than a single, all-encompassing model.

\section{Acknowledgments}
\label{sec:ack}

We thank Google Cloud for the generous TPU grant.

\bibliographystyle{splncs04}
\bibliography{main}

\end{document}